\documentclass[12pt,a4paper]{article}
\pdfoutput=1
\usepackage{arxiv}

\usepackage[backref=page, breaklinks, colorlinks=false]{hyperref}
\usepackage{amsmath}
\usepackage{amsfonts}
\usepackage{amssymb}
\usepackage{graphicx}
\usepackage{booktabs}
\usepackage[table,xcdraw]{xcolor}
\usepackage{subcaption}
\usepackage{glossaries}

\usepackage{tabularx}
\usepackage{rotating}
\newcolumntype{C}{>{\centering\arraybackslash}X}

\makeglossaries

\newacronym{nwafu}{NWAFU}{Northwest A\&F University}
\newacronym{cvat}{CVAT}{Computer Vision Annotation Tool}
\newacronym{leap}{LEAP}{LEAP Estimates Animal Poses}
\newacronym{tleap}{T-LEAP}{Temporal LEAP}
\newacronym{dleap}{D-LEAP}{Deeper LEAP}

\newacronym{cpm}{CPM}{Convolutional Pose Machines}

\newacronym{mse}{MSE}{Mean Squared Error}
\newacronym{rmse}{RMSE}{Root Mean Squared Error}
\newacronym{med}{MED}{Mean Euclidian Distance}
\newacronym{pck}{PCK}{Percentage of Correct Keypoints}
\newacronym{pckh}{PCKh}{Percentage of Correct Keypoints normalized to the head length}

\newacronym{gan}{GAN}{Generative Adversarial Neural network}
\title{T-LEAP: Occlusion-robust pose estimation of walking cows using temporal information}
\author{Helena Russello\\
	Farm Technology Group\\
	Wageningen University \& Research\\
	Wageningen, The Netherlands\\
	\texttt{helena@russello.dev} \\
	\And
	Rik van der Tol\\
	Farm Technology Group\\
	Wageningen University \& Research\\
	Wageningen, The Netherlands\\
	\texttt{rik.vandertol@wur.nl} \\
	\And
	Gert Kootstra\\
	Farm Technology Group\\
	Wageningen University \& Research\\
	Wageningen, The Netherlands\\
	\texttt{gert.kootstra@wur.nl} \\
}
\begin{document}

\maketitle

\begin{abstract} %
As herd size on dairy farms continues to increase, automatic health monitoring of cows is gaining in interest.
Lameness, a prevalent health disorder in dairy cows, is commonly detected by analyzing the gait of cows. 
A cow's gait can be tracked in videos using pose estimation models because models learn to automatically localize anatomical landmarks in images and videos.
Most animal pose estimation models are static, that is, videos are processed frame by frame and do not use any temporal information. 
In this work, a static deep-learning model for animal-pose-estimation was extended to a temporal model that includes information from past frames. 
We compared the performance of the static and temporal pose estimation models.
The data consisted of 1059 samples of 4 consecutive frames extracted from videos (30 fps) of 30 different dairy cows walking through an outdoor passageway. 
As farm environments are prone to occlusions, we tested the robustness of the static and temporal models by adding artificial occlusions to the videos.
The experiments showed that, on non-occluded data, both static and temporal approaches achieved a Percentage of Correct Keypoints (PCKh@0.2) of 99\%.
On occluded data, our temporal approach outperformed the static one by up to 32.9\%, suggesting that using temporal data was beneficial for pose estimation in environments prone to occlusions, such as dairy farms.
The generalization capabilities of the temporal model was evaluated by testing it on data containing unknown cows (cows not present in the training set).
The results showed that the average PCKh@0.2 was of 93.8\% on known cows and 87.6\% on unknown cows, indicating that the model was capable of generalizing well to new cows and that they could be easily fine-tuned to new herds.
Finally, we showed that with harder tasks, such as occlusions and unknown cows, a deeper architecture was more beneficial.

\end{abstract}

\section{Introduction}\label{sec:intro}
\begin{figure}[ht]
    \centering
    \subfloat{\includegraphics[width=0.33\linewidth]{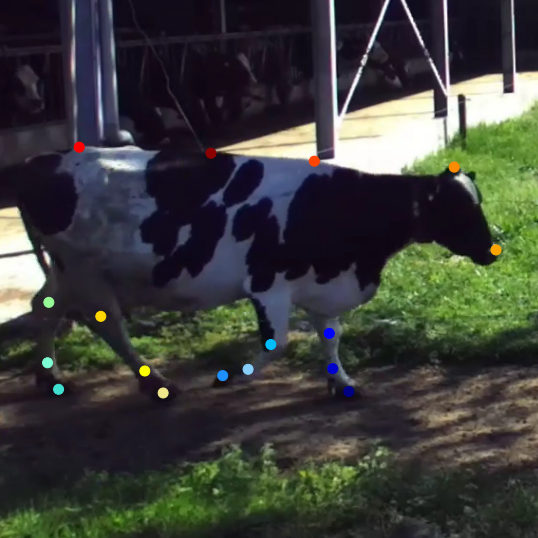}}
    \subfloat{\includegraphics[width=0.33\linewidth]{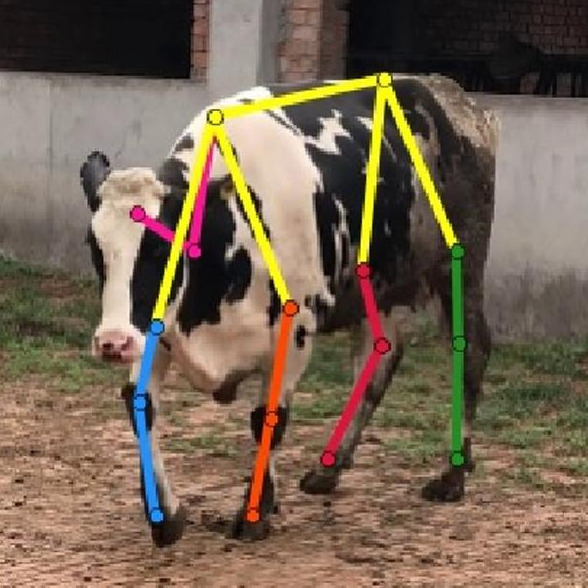}}
    \subfloat{\includegraphics[width=0.33\linewidth]{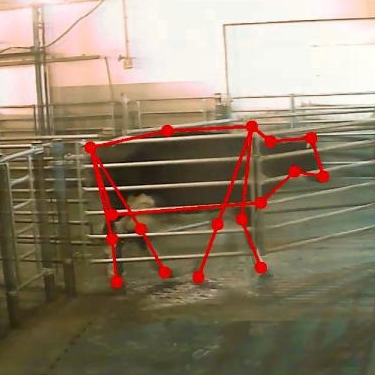}}
    \caption{Example of pose estimation on cows, from  this work (left), \cite{li2019deep} (center) and \cite{liu2020video} (right).
    }
    \label{fig:example-pose-est}
\end{figure}

According to the U.S. Department of Agriculture\footnote{https://www.progressivedairy.com/stats}, the average herd size on U.S. dairy farms was 297 in 2020, an increase of 42\% in 10 years. 
In eight states, the average herd size was even more than 1000 dairy cows. With such large herds, farmers need to rely on computer monitoring to maintain welfare and production levels.
Lameness is a significant welfare issue in dairy farms and is often characterized by an abnormal gait of the cow.
In bio-mechanical research, the gait is typically assessed by analyzing the kinematics of markers placed at anatomical landmarks~\cite{flower2005hoof, blackie2013associations}. 
However, physical markers are prone to dirt and don't scale well with large herds.
As an alternative, deep-learning based pose-estimation models can track markerless anatomical landmarks in images and videos (Figure ~\ref{fig:example-pose-est}).

Recently, a considerable body of literature has grown around the theme of markerless animal-pose-estimation. 
Pose-estimation on laboratory animals were tested on fruit flies~\cite{pereira2019fast, gunel2019deepfly3d, mathis2018deeplabcut}, mice~\cite{pereira2019fast, mathis2018deeplabcut}, locusts~\cite{graving2019deepposekit} and C. elegans~\cite{hebert2020wormpose, li2020deformation} recorded in controlled environments. 
As pose estimation datasets for animals are often limited or non-existent, studies used transfer learning~\cite{mathis2020imagenet}, cross-domain adaptation~\cite{cao2019cross} or synthetic data~\cite{li2020deformation, fangbemi2020zoobuilder} to estimate the pose of animals where little data was available.
Several studies applied existing human-pose-estimation models to quadrupeds such as pigs~\cite{khan2020bottom}, dogs~\cite{kearney2020rgbd} and cattle~\cite{li2019deep}.
Lastly, \cite{liu2020video} proposed a video analytic system based on DeepLabCut~\cite{mathis2018deeplabcut} to detect the shape of the cows' body as well as the legs and hoofs.

Most animal-pose-estimation models are static. 
That is, they consider the frames from videos as independent images and do not explicitly leverage the temporal information that is inherent to videos. 
Some impose temporal constraints~\cite{graving2019deepposekit, liu2020video} or hand-craft temporal features~\cite{liu2020video}, but they do not learn spatio-temporal features in an end-to-end fashion.
Moreover, animal-pose-estimation models are often evaluated with data recorded in controlled environments, such as laboratories. 
Farms, however, are highly uncontrolled environments, with varying light conditions and non-uniform background. 
Additionally, the animals that need to be observed by video are frequently occluded by objects on the farm or by other animals. 
These occlusions often lead to inaccurate pose estimations as existing static models solely rely on the current spatial information. 
Temporal models, however, also include information from the past frames and therefore have the potential to generate more accurate pose estimations when occlusions occur.
Until the work of~\cite{mathis2020imagenet}, animal-pose-estimation studies did not directly address out-of-domain robustness, that is, the ability to generalize to new (previously unseen) individuals and environments. 
Instead, only the within-domain robustness was addressed, that is, the performance on known individuals and backgrounds.
In the case of dairy farms, out-of-domain robustness is crucial as pose-estimation models need to be able to generalize to new cows and new farms. 
In fact, it would be time consuming and expensive to have to label data for each cow on each farm.

In this study, we focus on the pose estimation of cows in outdoor conditions. 
To deal with the complexity of the scenes, we propose two extensions to the \gls{leap}~\cite{pereira2019fast} animal-pose-estimation neural network: (1) a deeper architecture to handle the increased complexity of the data, and (2) a temporal architecture to leverage the temporal information of the videos to better deal with occlusions.
We perform three experiments on the proposed pose estimation models.
First, we compare the performance of the static and temporal models by testing them on non-altered video frames and on video frames altered with artificial occlusions that simulate a more challenging environment.
Second, we evaluate how well the proposed temporal model can generalise to new/unknown cows by testing the temporal pose estimation model on videos frames of known and unknown cows.
Third, we run the occlusion and generalisation experiments on a shallower architecture (the original depth of LEAP) and compare the performance with our deeper version.
\section{Materials and Methods}\label{sec:material}
This section describes the materials and methods.
Section~\ref{sec:data-acquisition} details the data acquisition. 
To meet different purposes, the collected data were allocated to two datasets, CoWalk-10 and CoWalk-30 as described in Section~\ref{sec:datasets}.
The data pre-processing is described in Section~\ref{sec:data-pre}.
The deep neural networks for pose estimation are presented in Section~\ref{sec:pose-estimation}. Finally, Section~\ref{sec:experiments} describes the experiments and evaluation methods.

\subsection{Data acquisition}\label{sec:data-acquisition}

\begin{figure}[ht]

            \subfloat[Black and white cow - Sunny weather]{%
             \includegraphics[width=\linewidth]{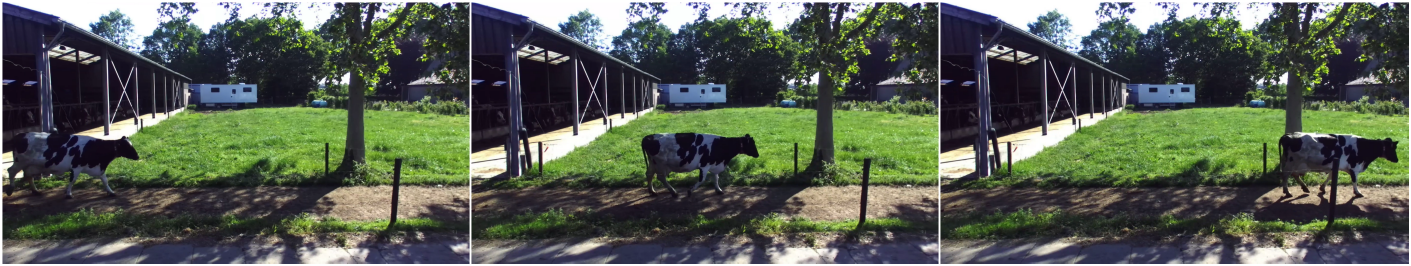}
            }\label{fig:example-frames1}
            \subfloat[Red and white cow - Sunny weather]{%
             \includegraphics[width=\linewidth]{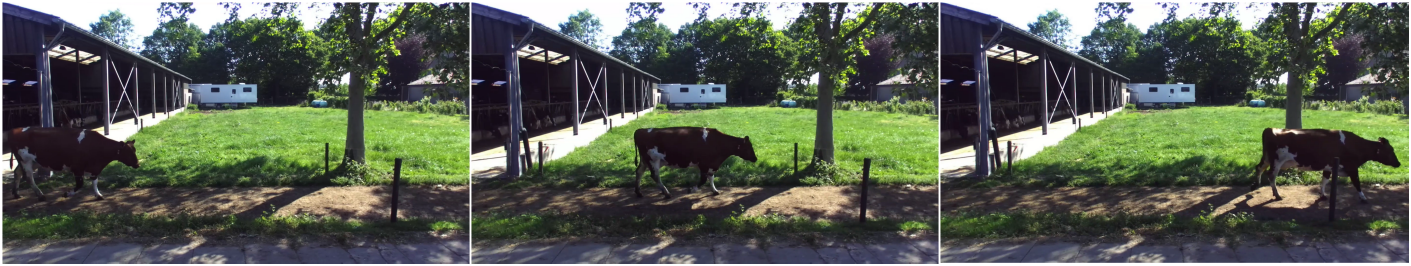}
            }\label{fig:example-frames2}
            \subfloat[Black and white cow - Cloudy weather]{%
             \includegraphics[width=\linewidth]{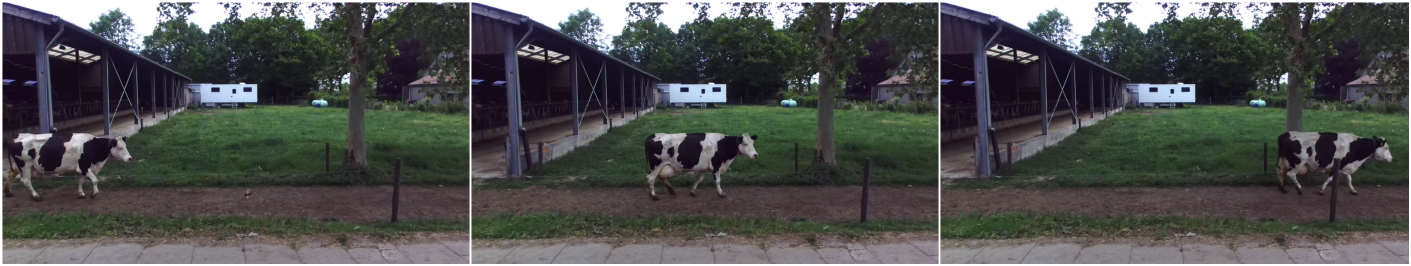}
            }\label{fig:example-frames3}
            \subfloat[Red and white cow - Cloudy weather]{%
             \includegraphics[width=\linewidth]{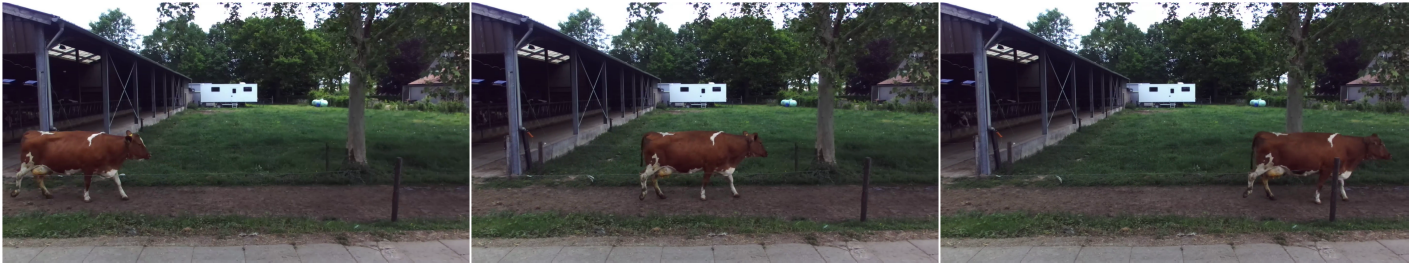}
            }\label{fig:example-frames4}
        
        \caption{Example frames extracted from videos with different cows and weather conditions.}
        \label{fig:example-frames}
        \end{figure}

The data consist of videos of walking Holstein-Frisian cows. 
The data were collected at a commercial dairy farm in Tilburg, The Netherlands, between 9 am and 4 pm,
on the 22nd and 27th of May and on the 3rd and 4th of June 2019. 
Seventy black-and-white and red-and-white Holstein-Friesian cows were filmed from the side with a ZED RGB-D stereo camera\footnote{\url{https://www.stereolabs.com/zed/}} (only RGB used), while walking from the barn to the pasture through a passageway (Figure~\ref{fig:example-frames}). 
The camera was placed 2 meters above the ground at 4.5 meters from the fence of walkway. 
The videos were filmed with a resolution of $1920 \times 1080$ pixels at 30 frames per second, with a spatial resolution in the field of view of approximately 0.2 pixels/mm. 
The 622 videos contained 226 frames on average. Hence, the videos were about 7.5 seconds long, which was the average time a cow needed to walk the visible part of passageway (9.5 meter). 
For creating the dataset, thirty videos of thirty different cows were selected according to the following criteria: a single cow was present in the passageway, the cow walked from the left to the right in the field-of-view and the cows walked without hesitance. 
The frames in which the body was not entirely visible were discarded.
In the selected videos, 20 cows were black-and-white and 10 were red-and-white (Figure~\ref{fig:example-frames}). 

\begin{figure}[ht]
    \centering
    \includegraphics[width=0.5\linewidth]{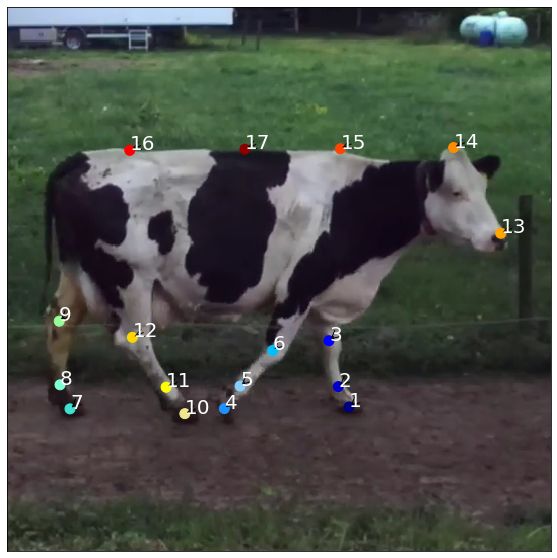}
    \caption{Anatomical landmarks as labeled in the CoWalk dataset. The keypoints are named as follows:
    1) LF hoof, 2) LF fetlock, 3) LF carpal, 
    4) RF hoof, 5) RF fetlock, 6) RF carpal,
    7) LH hoof, 8) LH fetlock, 9) LH carpal,            
    10) RH hoof, 11) RH fetlock, 12) RH carpal,
    13) Nose, 14) Forehead, 15) Withers, 16) Sacrum, 
    17) Caudal thoracic vertebrae. LF stands for Left-Fore, RF for Right-Fore, LH for Left-Hind and RH for Right-Hind.}
    \label{fig:example-labels1}
\end{figure}

The free and open-source Computer Vision Annotation Tool (CVAT)\footnote{\url{https://github.com/opencv/cvat}} was used to annotate the videos. 
In total, 4275 frames were annotated by one person, with an average of 138($\pm$30) frames per video. 
Seventeen anatomical landmarks on the cow's body were annotated (Figure~\ref{fig:example-labels1}). 
On the legs, the carpal or tarsal joints, fetlock joints and hoofs were labeled. 
On the back, the withers, caudal thoracic vertebrae and the sacrum were labeled. 
On the head the forehead and the nose were labeled. 
These anatomical landmarks were selected in consultation with experts in precision livestock farming. The localisation of these landmarks in the images is further referred to as keypoints.

\subsection{Datasets}\label{sec:datasets}
The annotated videos were split into samples of 4 consecutive frames with no overlap, resulting in 1059 samples.
Using these samples, two datasets were created: \textit{CoWalk-10}, containing training samples of 10 cows, and \textit{CoWalk-30}, containing training samples of 30 cows. 
The datasets are described hereafter and their number of samples is listed in Table~\ref{tab:numbers-dataset}.

\begin{table}[ht]
\centering
\caption{Number of samples in the training and test sets for of the CoWalk-10 and CoWalk-30 datasets.}
\label{tab:numbers-dataset}
\begin{tabular}{@{}lllll@{}}
\toprule
\textbf{Dataset}       & \textbf{Train} & \textbf{Test} (known cows) & \textbf{Test} (unknown cows) & \textbf{Total}  \\ \midrule
CoWalk-10  & 168   & 172               & 719                & 1059 \\ 
CoWalk-30     & 847  & 212              & N.A.                   & 1059 \\
\bottomrule
\end{tabular}
\end{table}

The \textit{CoWalk-10} dataset was used to evaluate our pose estimation methods' generalisation capacity to new individuals. To do so, we evaluated the methods' performance on known cows but unseen video frames and on unknown cows and unseen video frames. 
Following the same procedure as~\cite{mathis2020imagenet} we created a training set and two test sets: known-cows and unknown-cows test sets.
Out of the 30 videos, 10 videos were randomly selected for training.
From the 10 training cows, a random subset with 50\% of the samples was placed in the training set and the remaining 50\% in the known-cows test set. 
From the 20 remaining cows, all the samples were placed in the unknown-cows test set. 

The \textit{CoWalk-30} dataset was used to compare the performance of the static and temporal pose estimation methods. 
As the number of samples of the cowalk-10 training set was rather small, and as the generalisation to new cows was already tested with the cowalk-10 dataset, a dataset with more training samples and all 30 cows was created: the \textit{CoWalk-30} dataset. 
The dataset was divided following common practices in machine learning: a random subset of 80\% of the samples was placed in the training set and the remaining 20\% in the test set. 
Note that here, the same cows appear in the training and test sets. 
       
 \subsection{Data pre-processing}\label{sec:data-pre}
 \subsubsection{Preparing the image data}
\begin{figure}[!ht]
      \includegraphics[width=\linewidth]{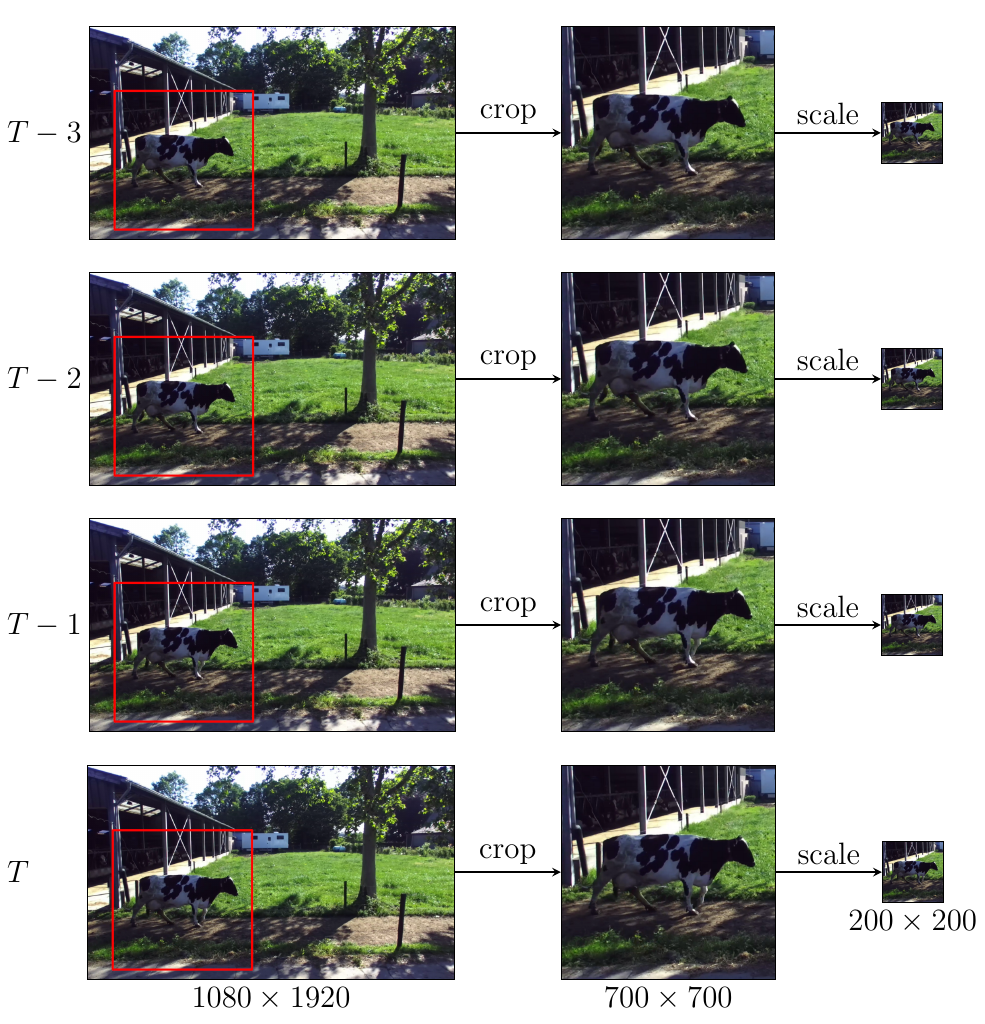}
 
        \caption{Workflow of the cropping and size reduction of a sample sequence. The bounding box is set on the first frame of the sequence ($T-3$) and used for the rest of the sequence.}
        \label{fig:example-cropping}
    \end{figure}

    To speed up the training process, we followed a similar procedure as~\cite{li2019deep} and cropped images around the cow's body before down-scaling them. 
   Our procedure is detailed bellow and pictured in Figure~\ref{fig:example-cropping}.
   The samples consist of sequences of 4 consecutive frames.
   The first frame of each sequence was cropped by placing a square bounding-box centered on the cow's body. 
   The location of the cow's body was retrieved using the location of annotated keypoints.
   The bounding box was extended by adding a 100-pixels margin to the front and hind side of the body, allowing the body of the cow to stay fully-visible throughout all frames of the sequence. As the bounding box was square, the height of the bounding box was equal to its width.
   The remaining frames of the sequence were cropped using the bounding box of the first frame.
   The size of frames was then reduced to $200\times200$ pixels, that is 3.5 times smaller on average. 
   
    To train the network more robustly and prevent over-fitting~\cite{simonyan2014very}, the number of training samples was artificially increased at each training epoch by performing rotation, brightness and contrast data augmentations using the \textit{OpenCV}\footnote{\url{https://opencv.org/}} python library.
    The rotation angle was between $[-10,10]$ degrees, the brightness noise was between $[-100,100]$ and the contrast gain was between $[-3.0, 3.0]$.
    The parameters for the rotation, brightness and contrast transformations were, for each sample, sampled randomly from a uniform distribution and were different at each epoch. 
    Note that no data augmentation was performed on the test set.
    
\subsubsection{Artificial occlusions}    
    
\begin{figure}[ht]
        \centering
        \subfloat[Hind legs]{%
             \includegraphics[width=0.24\linewidth]{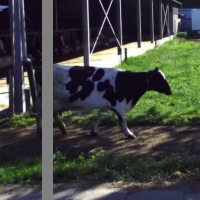}
            }
        \subfloat[Fore legs]{%
         \includegraphics[width=0.24\linewidth]{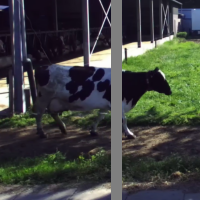}
        }
        \subfloat[Fore + Hind]{%
         \includegraphics[width=0.24\linewidth]{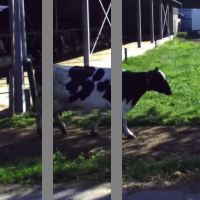}
        }
        \subfloat[Fore + Hind + Head]{%
         \includegraphics[width=0.24\linewidth]{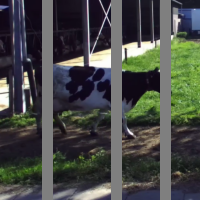}
        }
        \caption{Examples of occluded samples on the four occlusion test sets.}
        \label{fig:occlusions-examples}
    \end{figure}

    The samples of the CoWalk-30 test set were used to create four occlusion test sets in order to evaluate the pose-estimation methods in scenes where the cows' bodies were partially occluded.
    The occlusions consisted of grey vertical bars of 10 pixels width and were placed on all the frames, so that they were present during the whole sequence, simulating the presence of objects in the scene.
    The occlusions were placed in the video frames in four different ways: (a) around the hind legs, (b) around the front legs, (c) around the hind and front legs, and (d) around the hind legs, front legs and the head (Figure~\ref{fig:occlusions-examples}).
    The position of the occlusions was the same in all the video frames, and was determined by the mean position of the body part across all samples.
    
\subsection{Pose-estimation models}\label{sec:pose-estimation}
In this study, we adapted the LEAP model~\cite{pereira2019fast}, a popular animal-pose-estimation model.
The model attracted our attention because its neural-network architecture was smaller than other animal-pose-estimation models such as~\cite{mathis2018deeplabcut, graving2019deepposekit}, and provided a good trade-off between computational complexity and accuracy. Despite its smaller architecture, the authors of LEAP reported a good accuracy ($<2\%$ error rate) on two animal-pose-estimation datasets recorded in laboratory settings, that is, recorded with controlled light and uniform background.
Additionally, the simpler architecture of LEAP allowed more straightforward modifications to include temporal information.

We proposed two main modifications to LEAP: (1) the addition of convolutional layers to increase the receptive field and include more image context and (2) the modification of the LEAP model to a temporal model to estimate poses from sequences of images instead of single frames, further referred to as \acrfull{tleap}.
Both models are detailed hereafter\footnote{Code available at \url{https://github.com/hrussel/t-leap}}.

\subsubsection{LEAP}
\begin{figure}[ht]
    \centering
    \includegraphics[width=\linewidth]{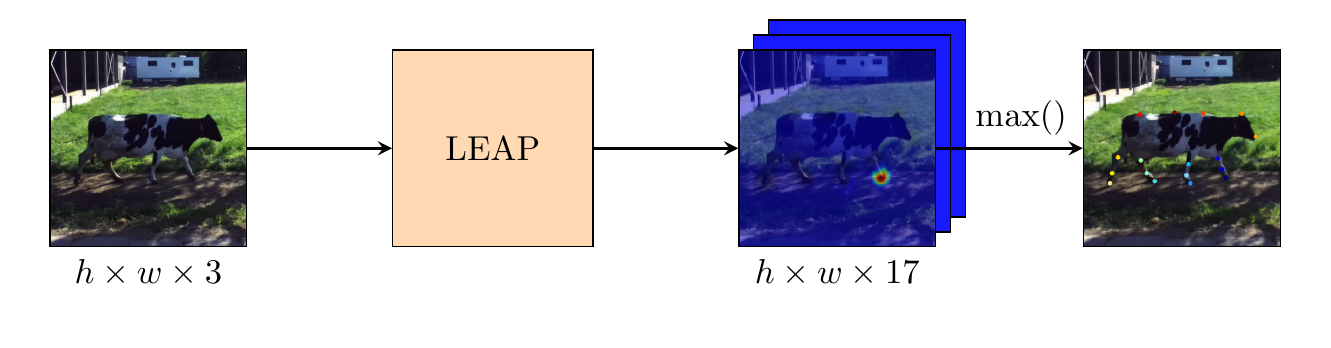}
    \caption{Workflow of the LEAP pose estimation.}
    \label{fig:leap-workflow}
\end{figure}

As shown in Figure~\ref{fig:leap-workflow}, LEAP takes an RGB image as input, and outputs one confidence map per keypoint.
A confidence map expresses, for each image coordinate, the probability that the given keypoint is located at that coordinate.
Hence, the location of a keypoint is determined by the highest value in its confidence map.
During training, the ground truth confidence-maps are generated per keypoint and per image. The generated confidence-maps consist of a 2D-Gaussian distribution centered around the ground-truth coordinates of the keypoint and $\sigma=5$.

\begin{figure}
    \centering
    \includegraphics[width=\linewidth]{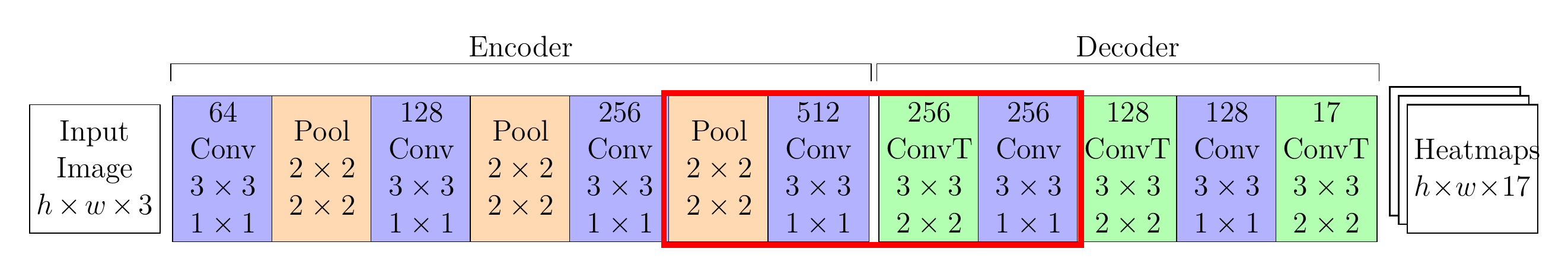}
    \caption{Architecture of the proposed LEAP model. 
    In each box, the text from top to bottom correspond to the output channels, operation, kernel size and stride. 
    The red rectangle indicates the layers that were added to the original architecture.
    Note that in the encoder, the convolutional blocks (in blue) consist of 3 convolutional layers, and in the decoder they consist of 2 convolutional layers.
    }
    \label{fig:leap-model}
\end{figure}

LEAP is a fully convolutional neural network and consists of an encoder and decoder part (detailed architecture in Figure~\ref{fig:leap-model}). The encoder extracts features from the images and the decoder up-samples the feature maps and outputs confidence maps of the same spatial dimension as the input.
LEAP was initially designed to work with data recorded in a controlled environment with uniform background. As opposed, our data were recorded outdoor, with varying daylight conditions and a non-uniform background. To make up for the added complexity of the data, and to increase the size of the receptive field in order to capture more image context, we increased the depth of the LEAP neural network by adding one convolutional group in the encoder and in the decoder.
The encoder now consists of 4 groups of 3 convolutional layers.
Between each group, a max-pooling layer reduces the size of the feature maps by half by using a kernel of size 2 with stride of 2 and no padding.
Each convolutional layer uses ReLU activation and batch-normalisation, and has a kernel of size 3 with stride of 1 and padding of 1. 
The decoder consists of 2 groups of 2 convolutional layers.
Before each group, a transposed-convolution layer doubles the size of the feature maps.
As in the encoder, the convolutional layers use ReLU activation and batch-normalisation.
The 2 transposed-convolution layers use a ReLU activation, and have a kernel of size 3 with stride of 2 and padding of 1. 
The last convolutional layer of the decoder is followed by a transposed-convolutional layer that uses a linear activation followed by a Softmax transformation, resulting in an output of 17 confidence maps, one per keypoint.

\subsubsection{Temporal LEAP}
\begin{figure}
    \centering
    \includegraphics[width=\linewidth]{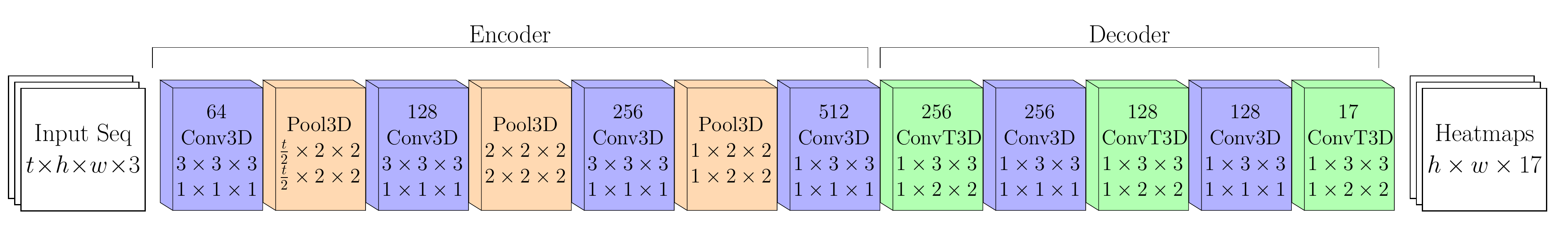}
    \caption{Architecture of the T-LEAP model. 
    In each box, the text from top to bottom correspond to the output channels, operation, kernel size and stride. 
    }
    \label{fig:tleap-model}
\end{figure}

In order to leverage the temporal information in videos, we proposed a temporal variant of the LEAP model that we called T-LEAP. 
The input did not consist of single frames anymore, but of a sequence of frames of length $T$.
The output, however, remained a single frame, and consisted of the confidence maps linked to the last frame $t=T$.
We kept the same architecture as the deeper LEAP, but modified all the convolution, pooling and transposed-convolution operations from 2D to 3D.
With a 3D convolution~\cite{tran2015learning}, the convolution operation was not only applied to the spatial dimension of the input, but also to the temporal dimension, allowing to learn spatio-temporal patterns from the data.
As the output concerned a single frame (the last frame in the sequence), the features of the temporal signal were aggregated during the max-pooling operation, and after the second max-pooling layer, the temporal signal was reduced to one.
The detailed architecture is shown in Figure~\ref{fig:tleap-model}.

\subsubsection{Training procedure}
    The models were implemented using the PyTorch~\cite{pytorch2019} deep-learning framework (version 1.6.0). The experiments were run on a high-performance computing cluster equipped with a Nvidia V100 GPU and the training progress was tracked with the \textit{Weights and Biases (W\&B)} platform~\cite{wandb}.
    The best hyper-parameters were found by performing a grid search using the sweep tool of W\&B.
    These optimized hyper-parameters consist of a batch size of 8, the AMSGrad optimizer~\cite{sashank2018on} with an initial learning rate of 0.001, and a learning-rate decay of 0.1 every 10 epochs. 
    The models were trained for 50 epochs, and were found to converge after 40 epochs.
    The loss function calculated the \acrfull{mse} per pixel per map between the predicted confidence maps and the ground-truth confidence maps from the samples in the batch.

\subsection{Experiments and evaluation}\label{sec:experiments}
    In this study, we ran three experiments:
    \begin{enumerate}
        
    \item The static and temporal models' performances was compared in their ability to deal with occlusion. 
    LEAP was trained with single video frames (T=1) and T-LEAP with sequences of 2 (T=2) and 4 (T=4) consecutive video frames. Both models were trained on the CoWalk-30 dataset and then tested on the non-occluded and occluded test data. The non-occluded data consisted of the unaltered CoWalk-30 test set. The occluded data consisted of artificial occlusions added to the CoWalk-30 test set. 
    
    \item The level of generalization of the T-LEAP model was studied.
    The T-LEAP model was trained on the CoWalk-10 dataset with samples of 2 consecutive frames (T=2) and the performance was evaluated on known and unknown cows. 
    By comparing the performance on a test set of known cows and a test set of unknown cows, we can test if the pose estimation model has learned a general representation of the posture of the cows, or if it over-fitted and learned a one-to-one mapping of the keypoints to the training cows.
    
    \item The influence of a deeper neural network on the pose-estimation performance was investigated. To do so, the performance of T-LEAP  with the same depth as the original LEAP architecture was compared against our proposed deeper T-LEAP architecture on the experiments of the CoWalk-30 and CoWalk-10 datasets.
    For CoWalk-30, both models were trained on the CoWalk-30 dataset and tested on frame sequences without occlusions and with 3 occlusions (Front legs, Hind legs and Head). 
    For CoWalk-10, both models were trained on the CoWalk-10 dataset and tested on frames sequences of known and unknown-cows.
    
    \end{enumerate}
\subsubsection{Evaluation metric}

    To evaluate the performance of the models, we used the \acrfull{pckh}~\cite{andriluka20142d}.
    The PCKh metric is commonly used in pose estimation~\cite{li2019deep, wei2016convolutional, newell2016stacked} and expresses the percentage of correct keypoints, where a predicted keypoint is considered correct if its distance to the ground-truth keypoint is smaller than a fraction of the head length. For instance, PCKh@0.5 is the percentage of keypoints within the threshold of half the head length. Let $h_i$ be the length of the head (calculated from the ground-truth) for data $i$, $\mathbf{p}_i$ the predicted position of the keypoint, $\mathbf{t}_i$ the ground-truth position of the keypoint and $\theta$ the proportional threshold, then the PCKh is defined as:
   $$
       \text{PCKh@}\theta = \frac{1}{N} \sum_{i=1}^{N}  \sigma (|| \mathbf{p}_{i} - \mathbf{t}_{i} || - h_i*\theta)
   $$
   where $\sigma(x) = 1$ when $x \leq 0$ and $\sigma(x) = 0$ otherwise.
   By using the head length as a threshold, the PCKh metric is independent of the size of the inputs and of the body of each animal. 
   Selecting the right threshold depends on the application of the pose estimation.
   Using a threshold of 0.5 would be too forgiving on the keypoints located on the leg. For instance, a right tarsal keypoint that is predicted on the left leg could be considered correct with a threshold of 0.5. 
   On the other hand, using a threshold of 0.1 or even 0 would be too restrictive and would consider too many keypoints incorrect. In fact, the task at hand is to estimate the location of keypoints, not to find its exact location at a pixel level.
   Hence, we consider the head threshold at 0.2 a good measure of accuracy for our application and it corresponds to $\sim10$cm.
To test the statistical significance of differences in performance, the Wilcoxon signed-rank test was used. As a level of significance, $p<0.05$ was used.
\section{Results}\label{sec:results}
The results of the experiments on occlusions, generalization and depth of the network are presented in the following subsection. 

\subsection{Dealing with occlusions: Static vs. Temporal (Experiment 1)}
\label{sec:results-static-temporal}

\begin{figure}[!ht]
    \centering
    \includegraphics[width=1.\linewidth]{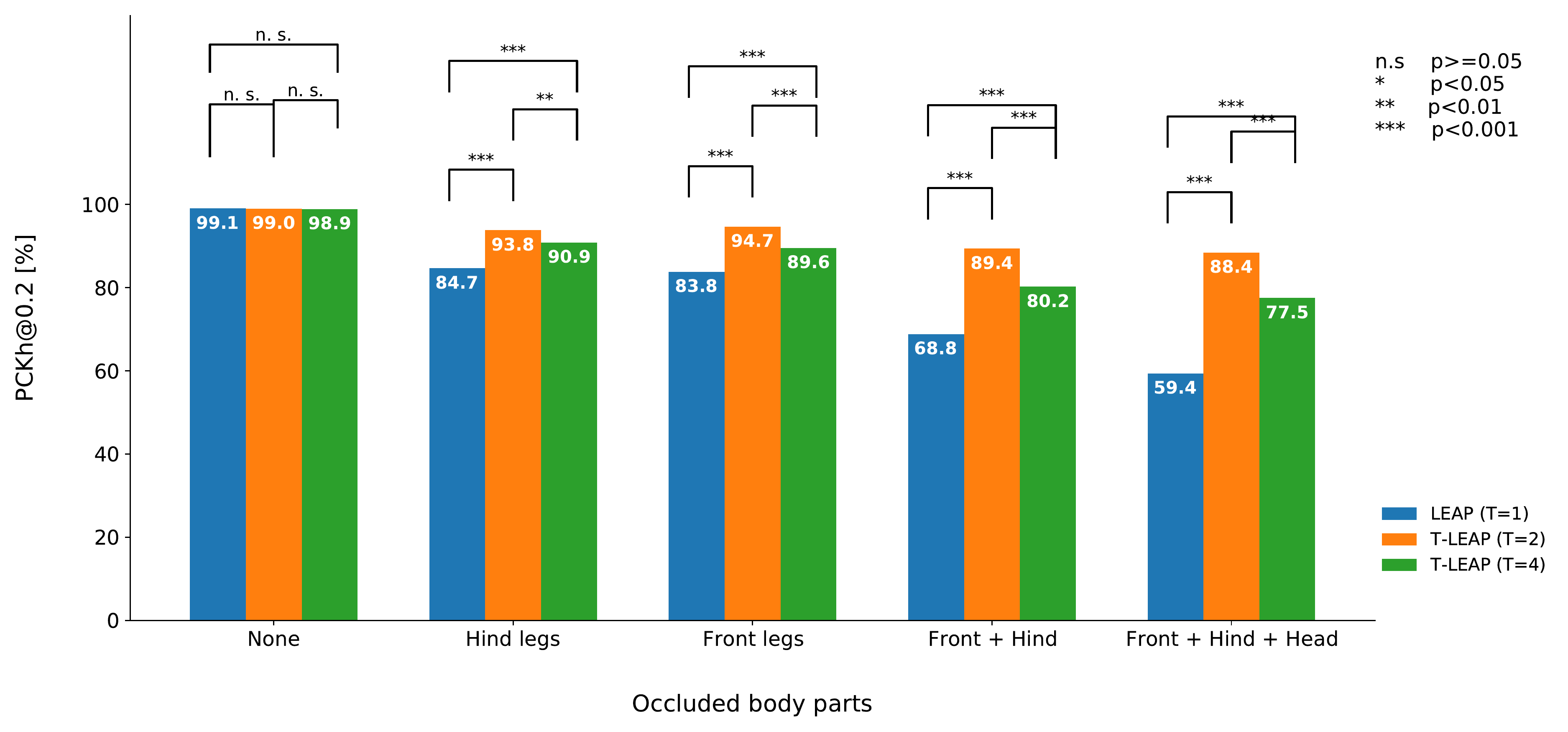}
    \caption{PCKh@0.2 of LEAP (T=1) and T-LEAP (T=2, T=4) on the CoWalk-30 test data with and without artificial occlusions. The occlusions were placed around the hind legs, fore legs and/or head.}
    \label{fig:pckh-occlusions}
\end{figure}

\begin{figure}[ht!]
\centering
\setlength\tabcolsep{2pt}%
\begin{tabularx}{0.75\textwidth}{@{}c*{3}{C}@{}}
~ & LEAP (T=1) & T-LEAP (T=2) & T-LEAP (T=4) \\
\rotatebox[origin=lB]{90}{No occlusion} &
   \includegraphics[ width=\linewidth, height=\linewidth, keepaspectratio]{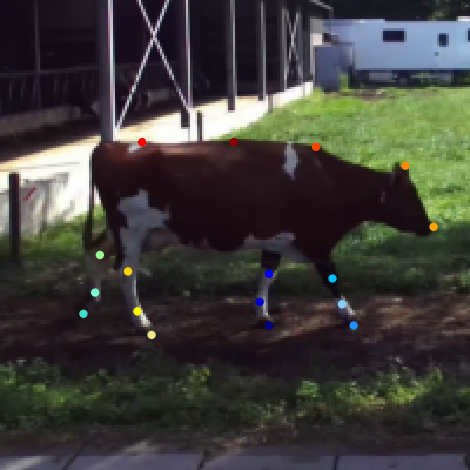} &
   \includegraphics[ width=\linewidth, height=\linewidth, keepaspectratio]{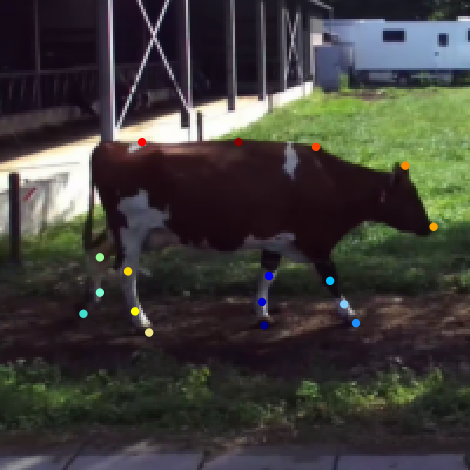} &
   \includegraphics[ width=\linewidth, height=\linewidth, keepaspectratio]{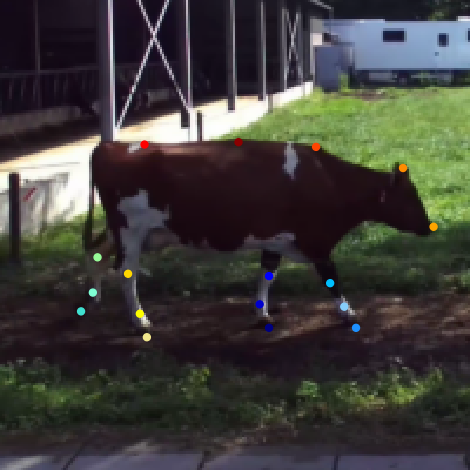}
 \\
 \rotatebox[origin=lB]{90}{Hind legs} &
   \includegraphics[ width=\linewidth, height=\linewidth, keepaspectratio]{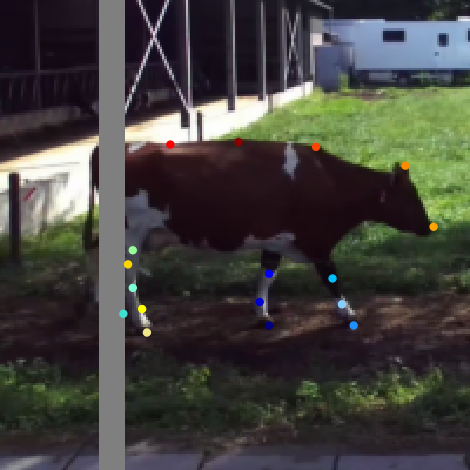}  &
   \includegraphics[ width=\linewidth, height=\linewidth, keepaspectratio]{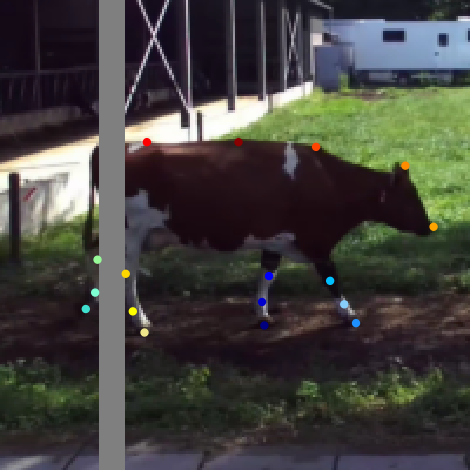}  &
   \includegraphics[ width=\linewidth, height=\linewidth, keepaspectratio]{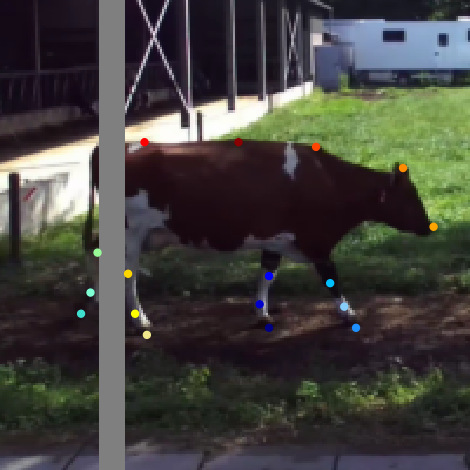} 
 \\
\rotatebox[origin=lB]{90}{Fore legs} &
   \includegraphics[ width=\linewidth, height=\linewidth, keepaspectratio]{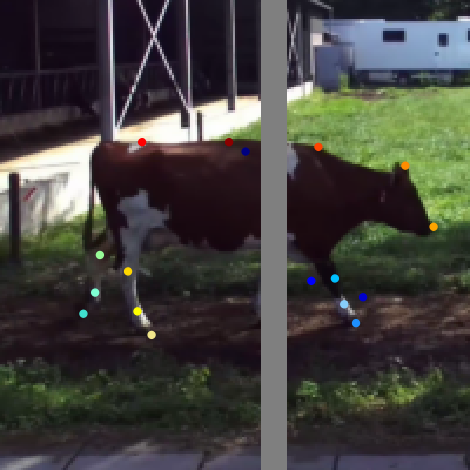}  &
   \includegraphics[ width=\linewidth, height=\linewidth, keepaspectratio]{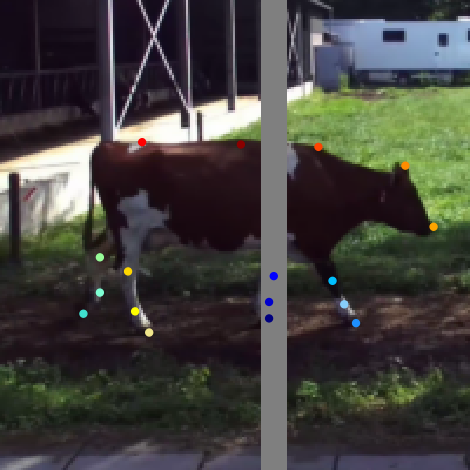}  &
   \includegraphics[ width=\linewidth, height=\linewidth, keepaspectratio]{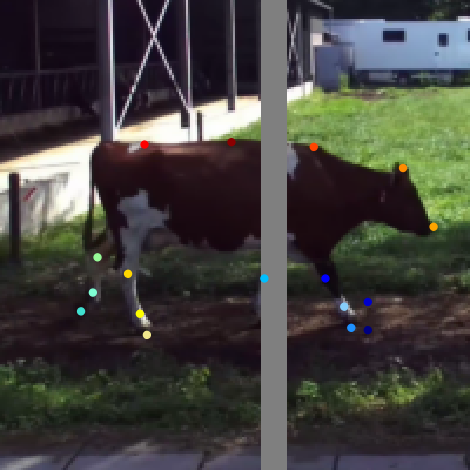} 
 \\
 \rotatebox[origin=lB]{90}{Fore, Hind} &
   \includegraphics[ width=\linewidth, height=\linewidth, keepaspectratio]{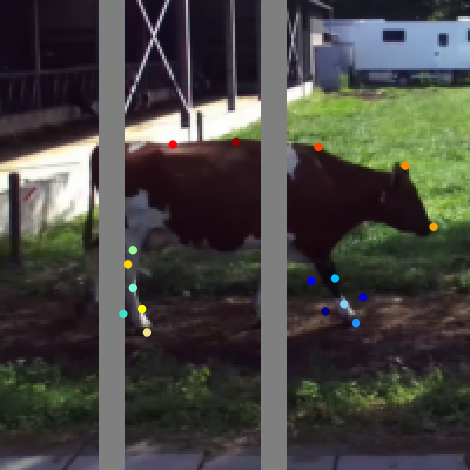}  &
   \includegraphics[ width=\linewidth, height=\linewidth, keepaspectratio]{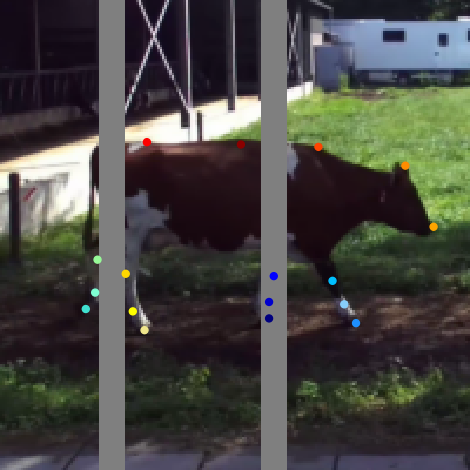}  &
   \includegraphics[ width=\linewidth, height=\linewidth, keepaspectratio]{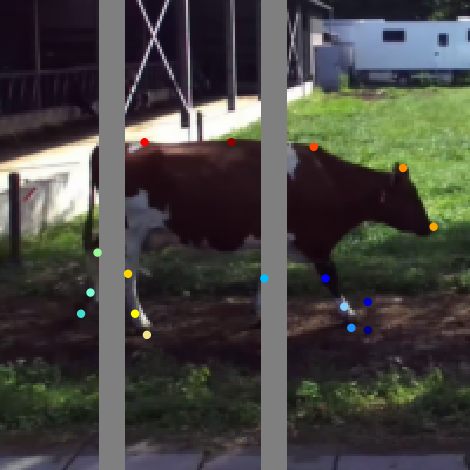} 
 \\
 \rotatebox[origin=lB]{90}{Fore, Hind, Head} &
   \includegraphics[ width=\linewidth, height=\linewidth, keepaspectratio]{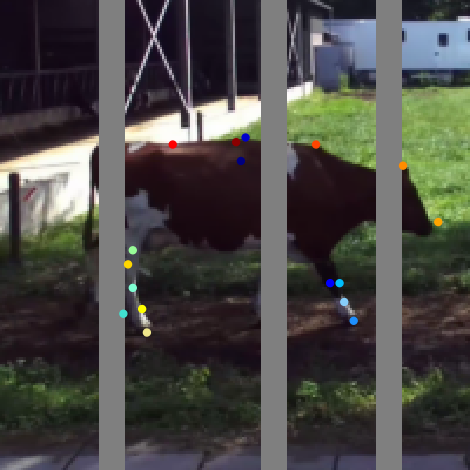}  &
   \includegraphics[ width=\linewidth, height=\linewidth, keepaspectratio]{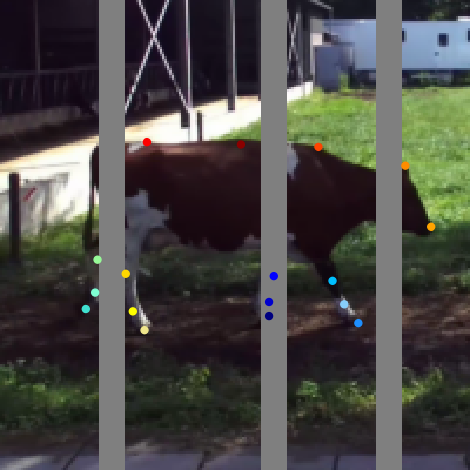}  &
   \includegraphics[ width=\linewidth, height=\linewidth, keepaspectratio]{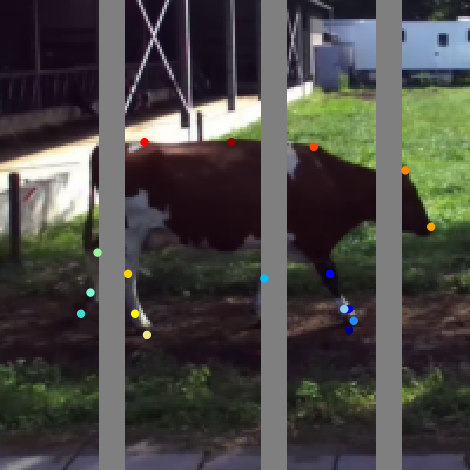} 

\end{tabularx}
\caption{Examples of the keypoints detection with LEAP and T-LEAP on one frame of the CoWalk-30 test data with and without artificial occlusions.} 
\label{fig:occlusions-results}
\end{figure}

The results presented in Figure~\ref{fig:pckh-occlusions} show the PCKh@0.2 of LEAP and T-LEAP on the non-occluded data and on the artificially occluded data as well as the p-value ranges of the statistical tests.
On the non-occluded data, all models achieved similar performance and could accurately detect keypoints with a mean PCKh@0.2 of 99\% for LEAP and T-LEAP. The differences between LEAP and T-LEAP with sequences of 2 frames (T=2), and between LEAP and T-LEAP with sequences of 4 frames (T=4) were not statistically significant ($p=0.356$ and $p=0.711$, respectively).
On the artificially occluded data, our temporal T-LEAP model outperformed the static LEAP model on all occlusions. 
The PCKh@0.2 was significantly higher for T-LEAP than for LEAP ($p=1\times10^{-20}$), and the gap became larger when more body parts were occluded. 
When adding occlusions to the hind legs, front legs, and the head, T-LEAP (T=2) maintained a PCKh@0.2 of 88.4\%, while LEAP only correctly detected 59.4\% of the keypoints.
Visual examples in Figure~\ref{fig:occlusions-results} further support that the temporal model could localise the occluded keypoints better.
In terms of sequence length, the occluded poses were significantly better detected with sequences of 2 frames than with sequences of 4 frames ($p = 5.42\times10^{-4}$).

\subsection{Generalization: Known-cows vs. Unknown-cows (Experiment 2)}
\label{sec:results-known-unknown}

\begin{figure}[!ht]
    \centering
    \includegraphics[width=\linewidth]{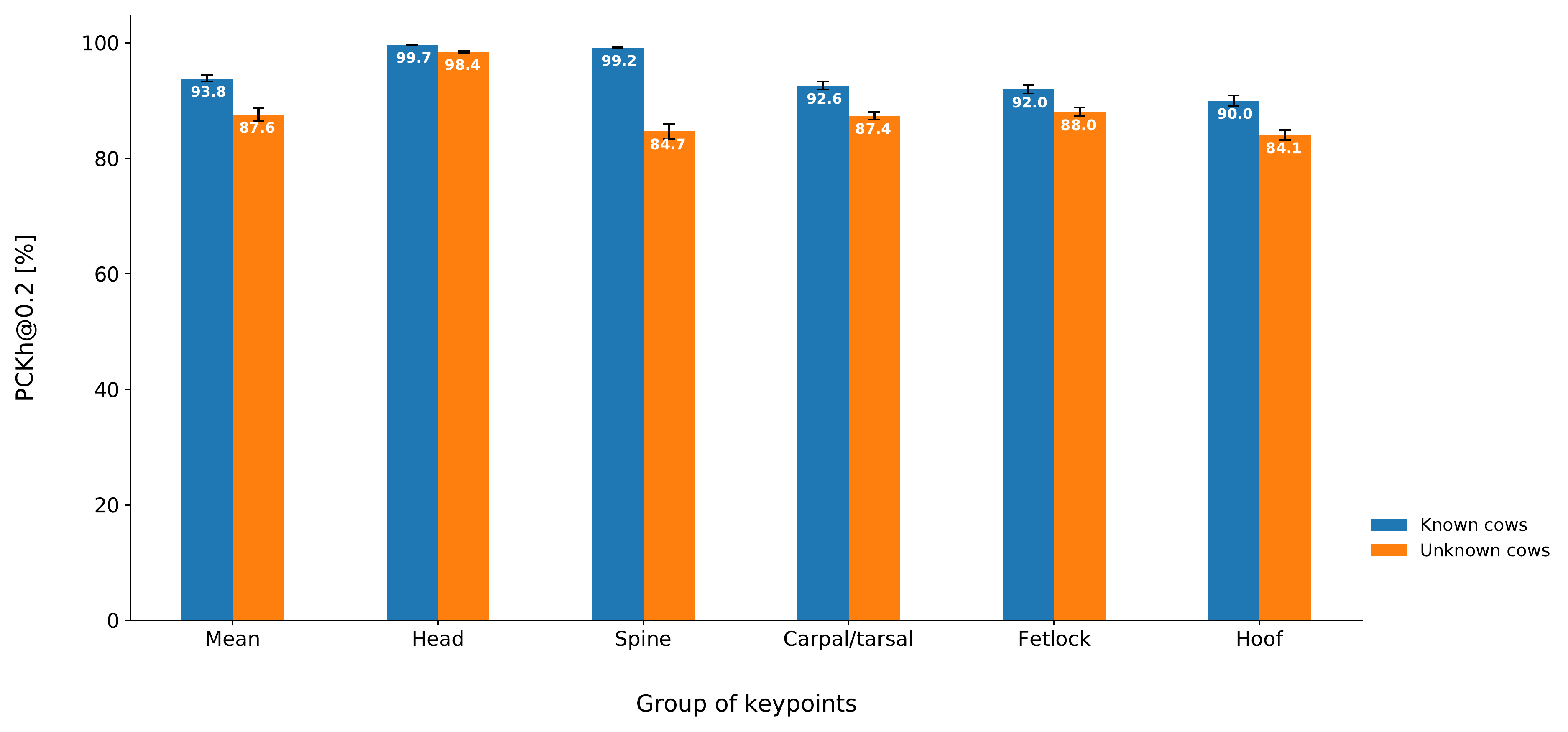}
    \caption{PCKh@0.2 per group of anatomical landmarks for T-LEAP (T=2) trained on CoWalk-10 and tested on the known-cows unknown-cows test sets.}
    \label{fig:cowalk-10-pckh}
\end{figure}

The PCKh@0.2 scores of the T-LEAP (T=2) model on the known and unknown cows test sets are shown in Figure~\ref{fig:cowalk-10-pckh}. Taking all body parts into account, the PCKh@0.2 drops from 93.8\% on known cows to 87.6\% on unknown cows. 
This 7\% performance decrease is the generalization gap, i.e., the difference between the performance of the model on training cows and its performance on unseen cows.
The head has a large PCKh@0.2, with 99.7\% on known cows and 98\% on unknown cows.
The keypoints on the spine, however, have a large PCKh@0.2 on known cows (99.2\%), but this drops to 84.7\% for the unknown cows.
The PCKh@0.2 of the keypoints on the legs drop from 92.6\% on known cows to 87.4\% on unknown cows for the carpal/tarsal keypoints, from 92\% to 88\% for the fetlock keypoints, and from 90\% to 84.1\% for the hoof keypoints.

\begin{figure}[!ht]
\centering
\setlength\tabcolsep{1pt}%
\begin{tabularx}{0.8\textwidth}{CCCCC}
& & Known cows & & \\

   \includegraphics[ width=\linewidth, keepaspectratio]{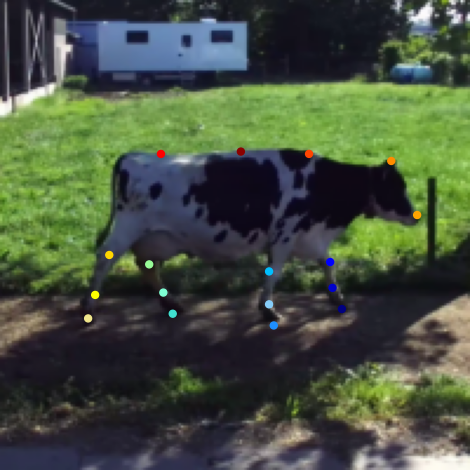} &
   \includegraphics[ width=\linewidth, keepaspectratio]{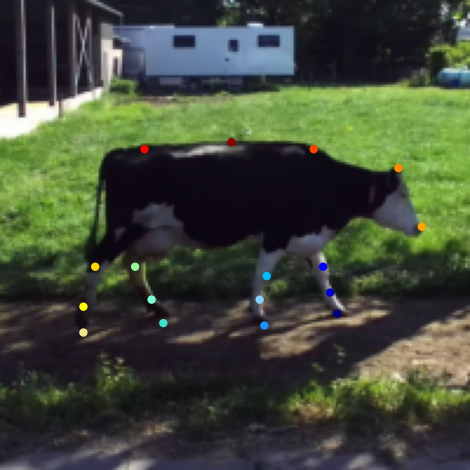} &
   \includegraphics[ width=\linewidth, keepaspectratio]{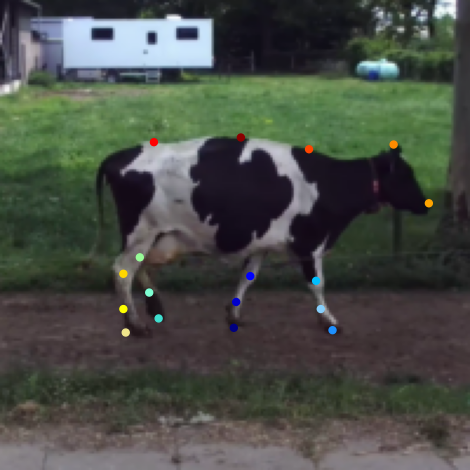} &
   \includegraphics[ width=\linewidth, keepaspectratio]{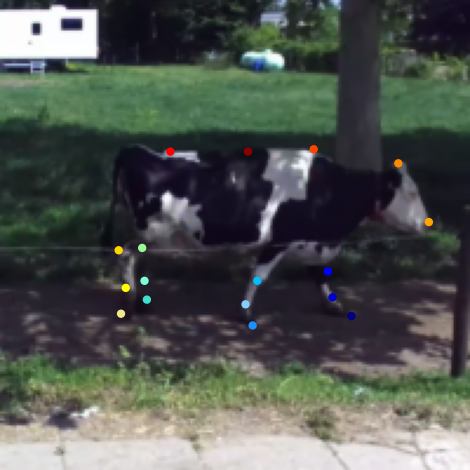} &
   \includegraphics[ width=\linewidth, keepaspectratio]{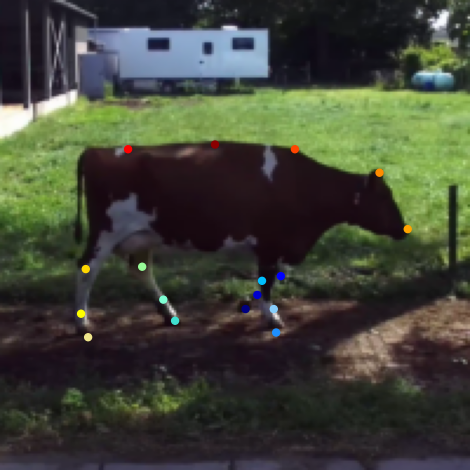} 
 \\
  \scriptsize{(1)}  & \scriptsize{(2)}  & \scriptsize{(3)}  & \scriptsize{(4)}  & \scriptsize{(5)}  \\
   \includegraphics[ width=\linewidth, keepaspectratio]{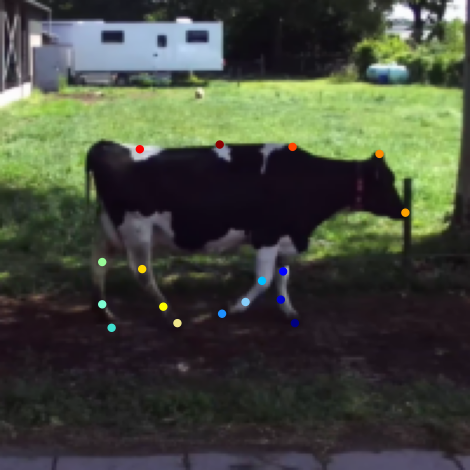} &
   \includegraphics[ width=\linewidth, keepaspectratio]{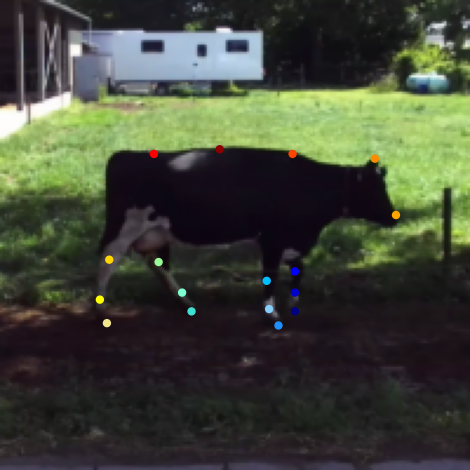} &
   \includegraphics[ width=\linewidth, keepaspectratio]{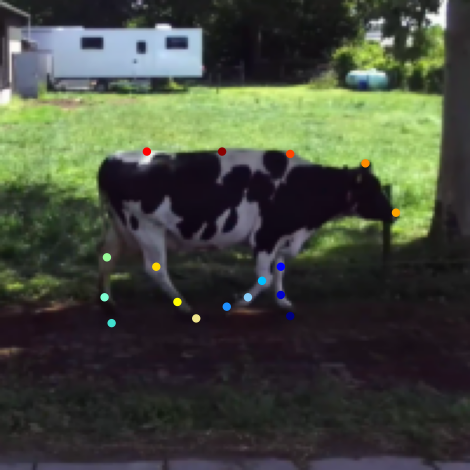} &
   \includegraphics[ width=\linewidth, keepaspectratio]{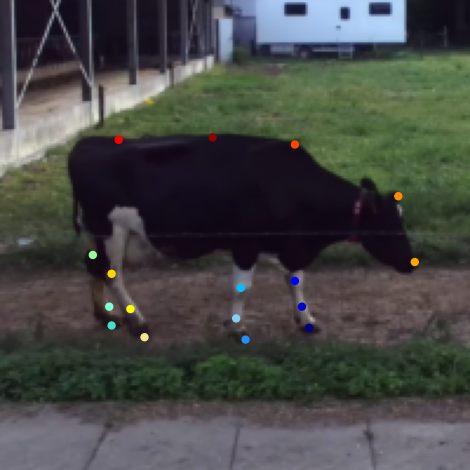} &
   \includegraphics[ width=\linewidth, keepaspectratio]{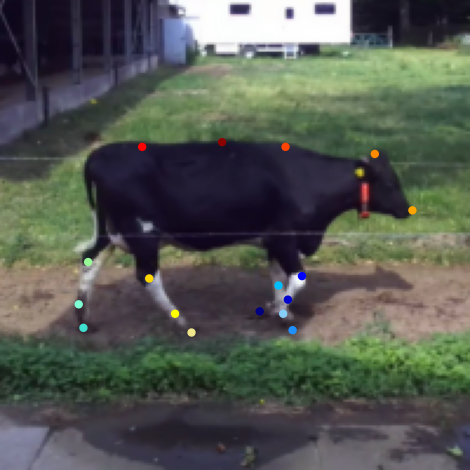} 
 \\
      \scriptsize{(6)}  & \scriptsize{(7)}  & \scriptsize{(8)}  & \scriptsize{(9)}  & \scriptsize{(10)}  \\
 & & Unknown cows & & \\
   \includegraphics[ width=\linewidth, keepaspectratio]{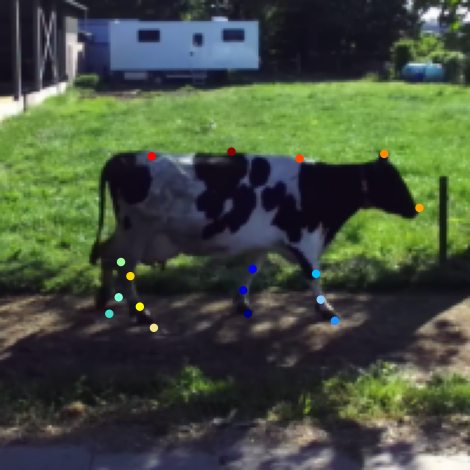} &
   \includegraphics[ width=\linewidth, keepaspectratio]{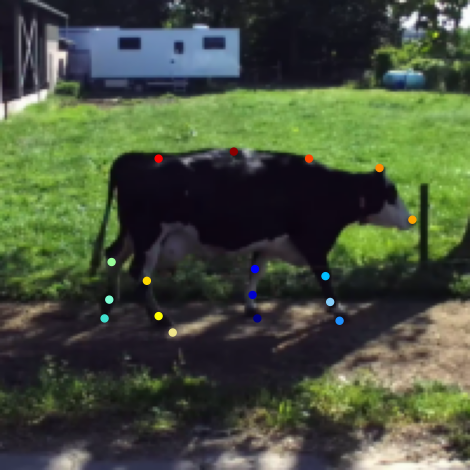} &
   \includegraphics[ width=\linewidth, keepaspectratio]{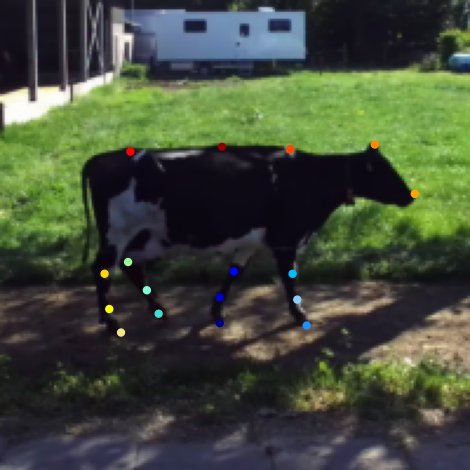} &
   \includegraphics[ width=\linewidth, keepaspectratio]{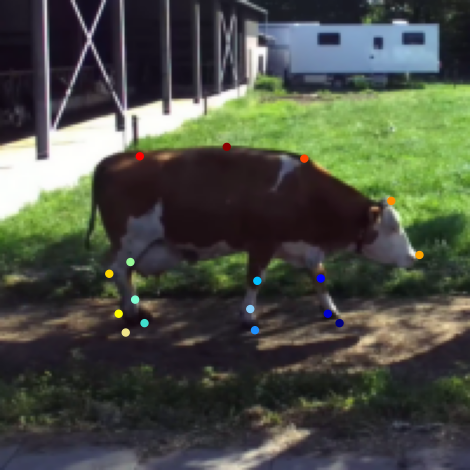} &
   \includegraphics[ width=\linewidth, keepaspectratio]{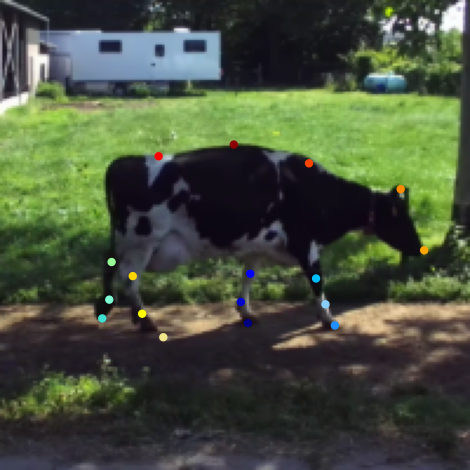} 
 \\
  \scriptsize{(11)}  & \scriptsize{(12)}  & \scriptsize{(13)}  & \scriptsize{(14)}  & \scriptsize{(15)}  \\
   \includegraphics[ width=\linewidth, keepaspectratio]{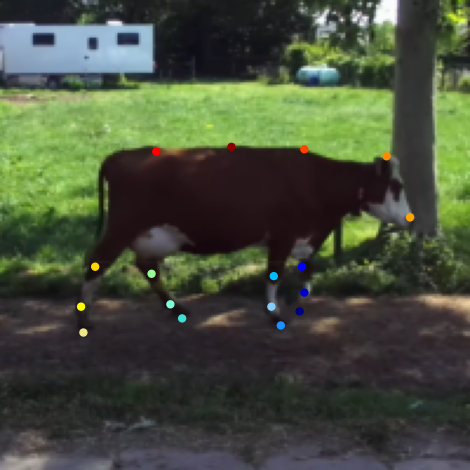} &
   \includegraphics[ width=\linewidth, keepaspectratio]{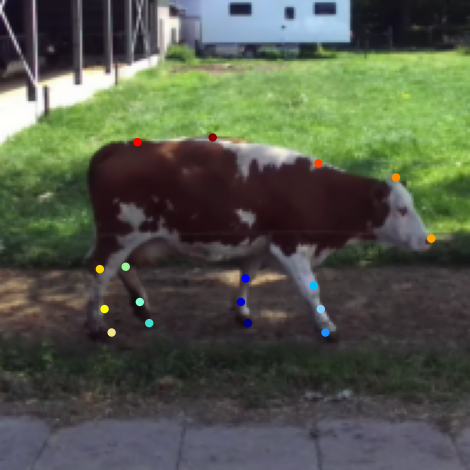} &
   \includegraphics[ width=\linewidth, keepaspectratio]{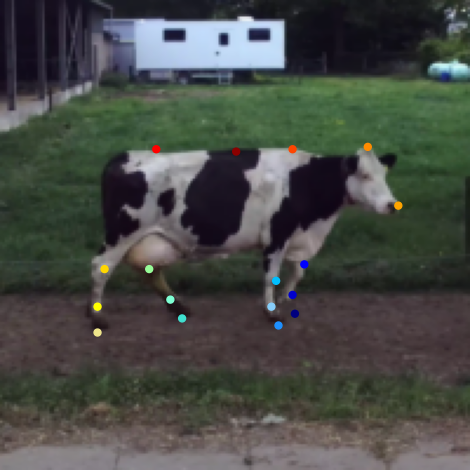} &
   \includegraphics[ width=\linewidth, keepaspectratio]{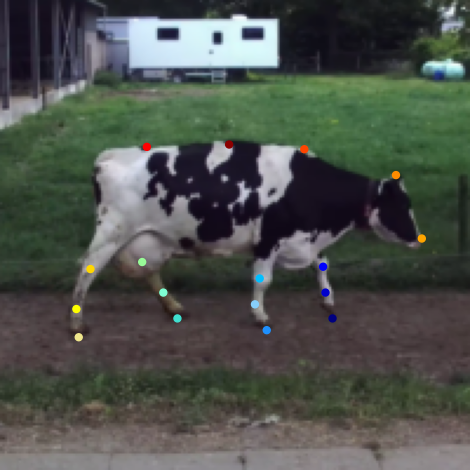} &
   \includegraphics[ width=\linewidth, keepaspectratio]{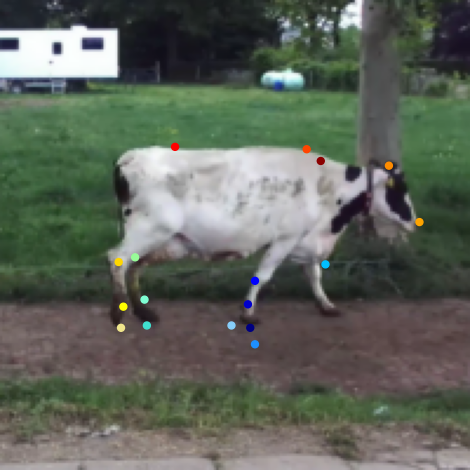} 
 \\
 \scriptsize{(16)}  & \scriptsize{(17)}  & \scriptsize{(18)}  & \scriptsize{(19)}  & \scriptsize{(20)}  \\
   \includegraphics[ width=\linewidth, keepaspectratio]{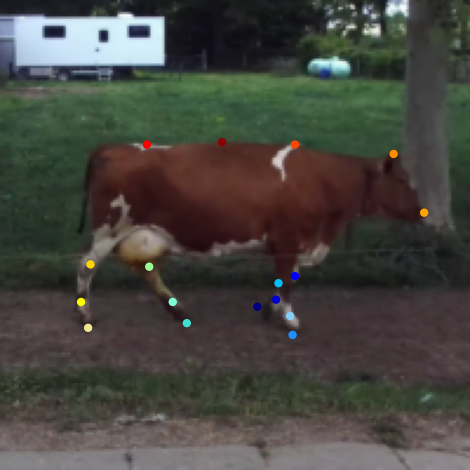} &
   \includegraphics[ width=\linewidth, keepaspectratio]{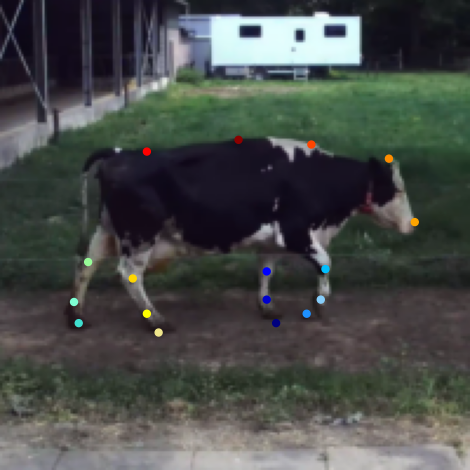} &
   \includegraphics[ width=\linewidth, keepaspectratio]{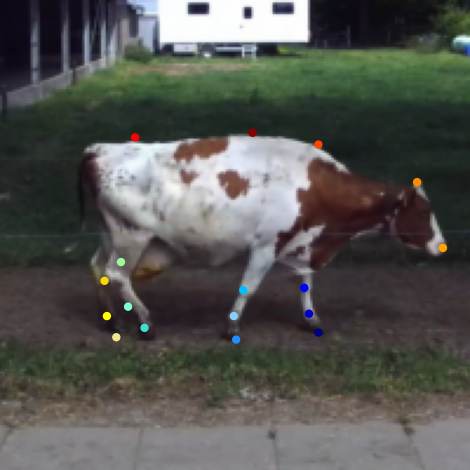} &
   \includegraphics[ width=\linewidth, keepaspectratio]{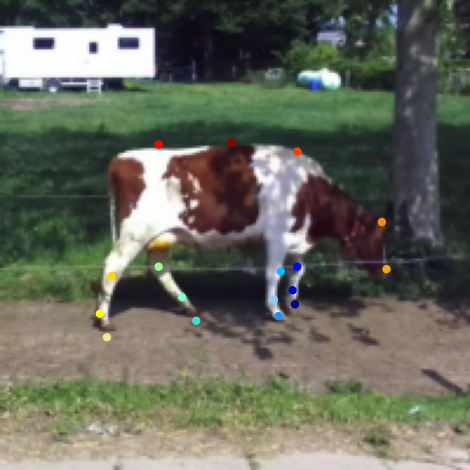} &
   \includegraphics[ width=\linewidth, keepaspectratio]{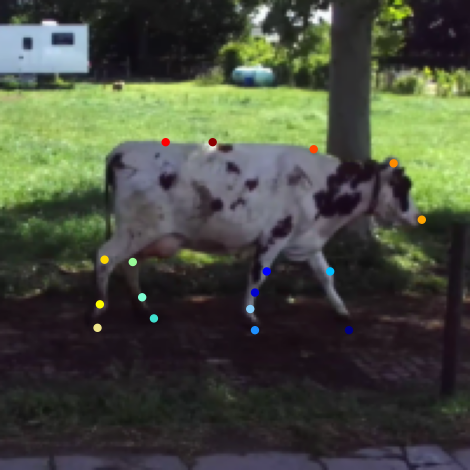} 
 \\
 \scriptsize{(21)}  & \scriptsize{(22)}  & \scriptsize{(23)}  & \scriptsize{(24)}  & \scriptsize{(25)}  \\
   \includegraphics[ width=\linewidth, keepaspectratio]{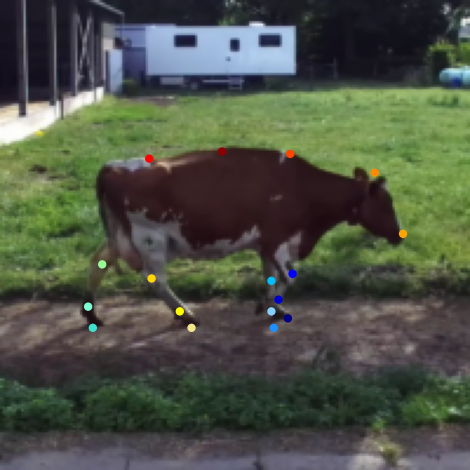} &
   \includegraphics[ width=\linewidth, keepaspectratio]{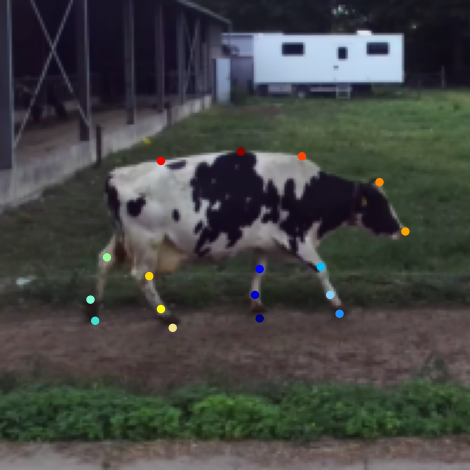} &
   \includegraphics[ width=\linewidth, keepaspectratio]{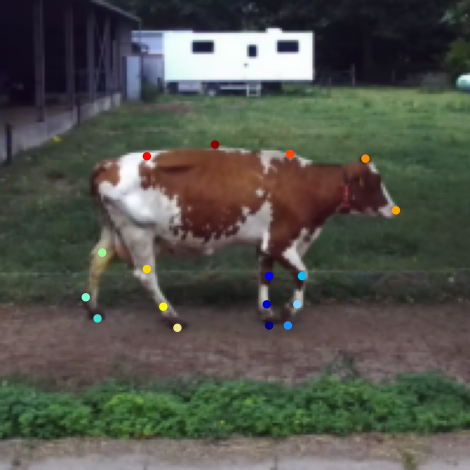} &
   \includegraphics[ width=\linewidth, keepaspectratio]{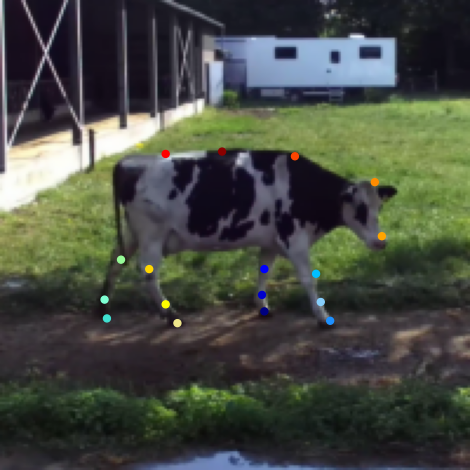} &
   \includegraphics[ width=\linewidth, keepaspectratio]{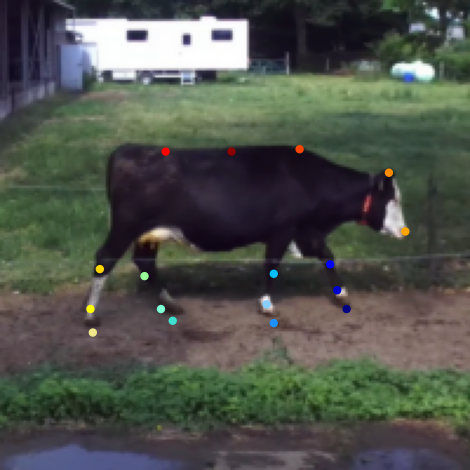} 
 \\
 \scriptsize{(26)}  & \scriptsize{(27)}  & \scriptsize{(28)}  & \scriptsize{(29)}  & \scriptsize{(30)}  \\
\end{tabularx}
\caption{Visual comparison of the results of T-LEAP (T=2) on known and unknown cows.} 
\label{fig:unknown-known-examples}
\end{figure}

Additionally, the test results were visually analysed, and a few examples of the known-cows and unknown-cows results are shown in  Figure~\ref{fig:unknown-known-examples}.
For the unknown-cows results, the visual analysis showed that the keypoints were well localized for the cows that had a coat pattern similar to that of the known cows (Figure~\ref{fig:unknown-known-examples}:(11)-(15)). However, for cows with a coat pattern deviating from all known-cows, the keypoints on the mid-spine and the lower legs were not well localized (Figure~\ref{fig:unknown-known-examples}:(20),(23),(25)).

\subsection{Depth of the network: Original vs. Deeper LEAP (Experiment 3)}
\label{sec:results-deeper-leap}

\begin{figure}[!ht]
    \centering
    \includegraphics[width=\linewidth]{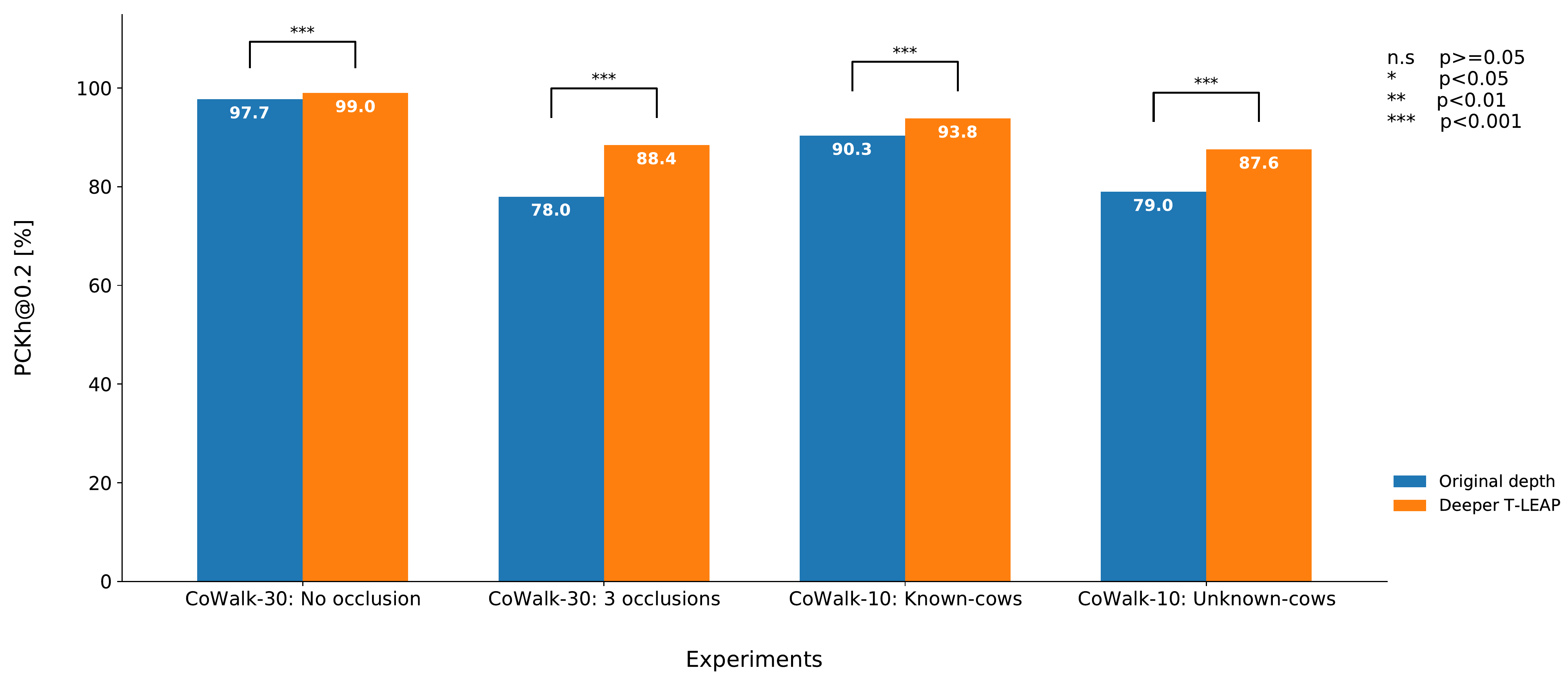}
    \caption{PCKh@0.2 of T-LEAP with original depth vs. our proposed deeper T-LEAP model on the most challenging experiments.}
    \label{fig:deeper-leap}
\end{figure}

 The results of the comparison between the T-LEAP (T=2) with the same depth as the original LEAP architecture, and our proposed deeper version are displayed in Figure~\ref{fig:deeper-leap}. 
 On both experiments with CoWalk-30 and with CoWalk-10, our deeper T-LEAP architecture performed significantly better.
 On experiments with CoWalk-30, using a deeper architecture improved the PCKh@0.2 by 1.3\% on data without occlusions, and improved by 11.7\% on data with three occlusions (Hind legs, Front legs and Head). 
 On experiments with CoWalk-10, the deeper architecture improved the PCKh@0.2 by 3.7\% on known-cows and by 9.8\% on unknown-cows.
  
\section{Discussion}
In the previous section, we performed three experiments on occlusion robustness (Experiment 1), generalisation to unknown-cows (Experiment 2) and depth of the network (Experiment 3) and presented the results. These results are discussed hereafter. 

In Experiment 1, the comparison of the static and temporal models on non-occluded and occluded data shows significant improvements of the temporal models on occlusions. 
These results indicate that the temporal model successfully learned spatio-temporal patterns from the data, which allows the model to provide more accurate pose estimation in the presence of occlusions.
A smaller temporal window of two consecutive video-frames is the most beneficial, suggesting that enough information is found in the immediate previous frame, and that longer-temporal patterns do not increase the PCKh of occluded poses for this data.
However, the needed temporal window probably relates to the size and severity of the occlusions. More specifically, the method may benefit from a longer temporal window in the presence of larger occlusions, as the body parts may be occluded during more frames.
The influence of the size of the occlusions on the temporal window should be investigated in future work.

An additional investigation could be to combine past and future frames to interpolate the location of keypoints in the current frame, instead of only using past frames to extrapolate the location of keypoints. 
This could give an advantage over only using two consecutive video-frames, as the future frame would provide more information on the occluded keypoints. Moreover, such an approach could also be more beneficial than using several past frames, as the data would be exactly one frame away from the analysed frame.

Experiment 2, the generalization experiment, showed that the performance of the pose-estimation model is high for known cows and for most unknown cows. 
However, the performance drops for unknown cows with a distinct coat pattern that was not represented in the training set, as seen with cows that have a mostly-white coat (Figure~\ref{fig:unknown-known-examples}:(20),(23),(25)).
These observations indicate the model is capable of generalizing to new cows if their coats are similar to the known cows. 
As such, we expect the model to generalize well to breeds that have uniform coats, such as Jersey cows.
For more heterogeneous breeds, a representative sample of the coat patterns should be present in the training set to allow for good generalization within and across herds.

In this experiment, only 10 cows were used for training out of the 30 annotated videos.
Adding more cows to the dataset would likely increase the robustness of the model on unseen coat colors and patterns, albeit increasing the annotation costs.

Future work could also investigate the use of a \acrfull{gan}, such as Cycle GAN~\cite{zhu2017unpaired}, to automatically generate new coat patterns and train the pose-estimation model to be more robust to differences in coat patterns.

In Experiment 3, all tests showed a significantly improved PCKh of our proposed deeper T-LEAP compared to the original architecture depth, and the improvement gap was especially larger for the tests with most challenging conditions including occlusions or unknown cows.
This indicates that the additional model parameters in the deeper architecture allow the network to deal with more complex data, whereas the shallower network was only able to deal with the less complex conditions.
The increased depth might help the pose estimation in two ways: (1) through the additional convolutional and pooling layers, neurons in the final layer of the encoder have a larger receptive field allowing to capture more spatial context, and (2) the additional layers allow the extraction of more complex spatio-temporal features. 

The data used in this study were recorded in realistic outdoor conditions, that is, with varying light, complex background and a fence partially occluding the cows. Moreover, we included artificial occlusions to resemble more challenging real-world situations. 
However, the postures were standardized as the cows walked in a straight line, and we selected videos containing only single cows. 
To allow the use of pose estimation in less constrained situations with cameras at different positions in the barn or in the field, future work should focus on even more challenging conditions including, for instance, different postures, multiple cows, and various barn elements.
Such added complexity might require more complex model architecture exploiting temporal and spatial information to a larger extend.

\cite{li2019deep} applied pose estimation on images of dairy and beef cattle.
Using their dataset, they trained three human-pose-estimation neural networks, namely Stacked-hourglass~\cite{newell2016stacked}, Convolutional-pose-machines~\cite{wei2016convolutional} and Convolutional-heatmap-regression~\cite{bulat2016human} to estimate the pose of cows.
A direct comparison of their performance against ours is not straightforward, as their data consisted of independent images of dairy and beef cattle in various postures taken from multiple viewpoints, whereas ours consisted of videos of only dairy cattle in standardized postures taken from a single viewpoint (side view), which is common in kinematic analysis~\cite{flower2005hoof, blackie2013associations}.
Their best model, the Stacked hourglass network, achieved a mean PCKh@0.5 of 90.39\%. The best detected body part was the head with a PCKh@0.5 of 97.15\%, and the hardest was the hoofs with a PCKh@0.5 of 83.90\%.
This is on par with our findings in Experiment 2, as we showed that for known cows, the head was the best detected, with a PCKh@0.2 of 99.7\%, and that the hoofs had the lowest PCKh@0.2 of 90\%.
Note that \cite{li2019deep} allowed a more lenient evaluation metric, as they set the threshold of the PCKh to 0.5, where we set it to 0.2. 
For reference, on Holstein-Frisian cows, a threshold of 0.2 of the head-size corresponds to approximately 10cm in real-world coordinates, whereas a threshold of 0.5 corresponds to approximately 25cm.
A larger threshold may be suitable for applications such as activity recognition, where the activity of the subject is inferred from the pose (e.g. standing or lying), but will not be suitable for gait analysis, where the movement of the different body parts need to be analyzed in detail.

\cite{liu2020video} proposed a system based on DeepLabCut~\cite{mathis2018deeplabcut} to extract the body shape and legs location of cows in videos. 
They trained one DeepLabCut neural networks on RGB images, and another one on "temporal difference", that is, the pixel-value difference between two consecutive frames. 
In a post-processing step, physical constraints were applied, as well as a temporal median filter to smooth the predictions.
In accordance with this study and with \cite{li2019deep}, they found that the keypoints on the legs and hoofs were the hardest to detect. 
Further studies should therefore focus on improving the detection accuracy of leg keypoints.
Their data consisted of videos of cows walking laterally through a pathway inside the barn. The fences of the pathway created challenging partial occlusions, but the effect to the occlusions was not explicitly studied.
Hand-crafted temporal features were included with the "temporal difference" and the median filter, however the approach did not seem to benefit from the "temporal difference" as that model was less temporally-consistent than the one trained with only (static) RGB data.
Furthermore, their approach relied on hand-crafted pre- and post-processing steps, and separate training of two neural networks, whereas our approach could be fully-trained end-to-end and was not limited to a sequence length.
Their post-processing module added physical constraints to the keypoints and can estimate the pose of multiple animals in the image. 
This post-processing approach could potentially be applied to any pose estimation models, and in combination with our T-LEAP, this could provide even better keypoint estimates. 

To the best of our knowledge, our proposed model is the first animal-pose-estimation model to learn spatio-temporal features in an "end-to-end" fashion. 
Although other existing animal-pose-estimation models have different neural-network architectures, they all consist of 2D convolutional neural networks. 
Therefore, our approach to use 3D convolutions could be applied to other pose-estimation models to compute spatio-temporal features. It is expected that by doing so, other animal-pose-estimation models will also benefit from temporal information when performing video-analysis of motion of body parts in occluded and challenging situations.

The proposed pose-estimation methods form an important foundation for automatic gait analysis. 
In future work, the pose estimation should be connected to biomechanical methods or morphometrics to compute gait features such as stride length, step symmetry and velocity (relative to body dimensions) to perform a kinematic analysis~\cite{flower2005hoof, blackie2013associations}, or to provide a gait score~\cite{wu2020lameness, gardenier2018object}. 

\section{Conclusion}\label{sec:conclusion}

In this study, we built upon the LEAP~\cite{pereira2019fast} model and implemented a (deeper) static and a temporal neural network for pose estimation of walking cows in videos.
The comparison of the static and temporal models on non-occluded and occluded data showed that the temporal models performed up to 32.8\% better than the static approach in presence of occlusions, and that a smaller temporal window of 2 consecutive video-frames was the most beneficial. 
The appropriate size of the temporal window, however, most likely depends on the severity of the occlusions.
The temporal model generalised well to unknown cows that were not seen in the training set with a generalisation gap of only 7\%.
For unknown cows that had a distinct coat pattern, the detection of the spine and hoof keypoints dropped, stressing the importance of a diverse training set.  
Finally, we showed that pose estimation on the most challenging conditions such as occlusions or unknown cows benefited the most from the additional parameters of the deeper architecture. 
Further investigation is required to evaluate the benefits and limitations of temporal pose estimation on dairy cows in challenging barn conditions by, for instance, including more complex postures and multiple animals in the field of view.
Another direction for future research might explore the use of the trajectories of anatomical landmarks detected with pose estimation to analyse the gait of dairy cows.

\section*{Acknowledgements}
This publication is part of the project Deep Learning for Human and Animal Health (with project number EDL P16-25-P5) of the research program Efficient Deep Learning (\url{https://efficientdeeplearning.nl}) which is (partly) financed by the Dutch Research Council (NWO).

\clearpage
\bibliographystyle{unsrt}  
\bibliography{sources}	
\end{document}